\PassOptionsToPackage{table}{xcolor}
\documentclass[10pt,twocolumn,letterpaper]{article}

\usepackage[pagenumbers]{cvpr}

\usepackage{cuted} %

\usepackage{currfile} %

\usepackage{caption} %

\usepackage[T1]{fontenc}
\usepackage{microtype}
\usepackage{graphicx}
\usepackage{booktabs}
\usepackage{colortbl}
\definecolor{tableheadgray}{gray}{0.9}
\newcommand{\tablehead}{\rowcolor{tableheadgray}}
\definecolor{rowgray}{gray}{0.96}
\newcommand{\headcell}[3]{\multicolumn{#1}{>{\columncolor{tableheadgray}}#2}{#3}}
\newcommand{\zebra}{\rowcolors{3}{white}{rowgray}}
\usepackage{amsmath}
\usepackage[most]{tcolorbox}

\usepackage{xr-hyper}
\definecolor{cvprblue}{rgb}{0.21,0.49,0.74}
\usepackage[pagebackref,breaklinks,colorlinks,allcolors=cvprblue]{hyperref}
\newif\ifwithsupplement
\withsupplementtrue
\ifwithsupplement\else
\fi
\makeatletter
\newcommand{\suppaliases}{\let\supp@newlabel\newlabel
  \def\newlabel##1##2{\supp@newlabel{##1}{##2}\supp@newlabel{S-##1}{##2}}}
\makeatother

\def\confName{CVPR}
\def\confYear{2027}

\newcommand{\notmeasured}[1]{\textcolor{red}{#1}}
\definecolor{holgergreen}{rgb}{0.0,0.45,0.0}
\newif\ifholgercolor
\holgercolorfalse
\newcommand{\holger}[1]{#1}
\definecolor{revgreen}{RGB}{0,110,40}

\definecolor{revblue}{RGB}{0,90,220}

\definecolor{revpurple}{RGB}{140,40,170}

\newcommand{\supp}[1]{Supp.~\ref{S-#1}}
\makeatletter
\newcommand{\setdefaultcolor}[1]{\color{#1}\let\default@color\current@color\let\XC@default@color\XC@current@color}
\makeatother

\newtcolorbox{findingbox}{colback=cvprblue!6, colframe=cvprblue, boxrule=0.6pt, arc=1pt, left=4pt, right=4pt,
                          top=3pt, bottom=3pt}

\definecolor{codecorange}{RGB}{217,130,43}
\usepackage{pifont}
\definecolor{relgreen}{RGB}{30,140,60}\definecolor{relred}{RGB}{200,40,40}
\newcommand{\cmark}{\textcolor{relgreen}{\ding{51}}}
\newcommand{\xmark}{\textcolor{relred}{\ding{55}}}
\newcommand{\passc}[1]{\textcolor{relgreen}{#1}}\newcommand{\failc}[1]{\textcolor{relred}{#1}}
\definecolor{renoblue}{RGB}{47,111,168}
\makeatletter
\renewcommand\paragraph{\@startsection{paragraph}{4}{\z@}{0.6ex \@plus .3ex \@minus .2ex}{-0.8em}{\normalfont\normalsize\bfseries}}
\makeatother
\newcommand{\renoname}{RENO}
\newcommand{\codecname}{\textcolor{codecorange}{\textbf{Spindle}}}
\title{The \texorpdfstring{\textcolor{codecorange}{Spindle}}{Spindle} Lidar Codec: Point Cloud Compression for Spinning Lidars}

\author{Holger Caesar\\
Intelligent Vehicles, Delft University of Technology\\
{\tt\small h.caesar@tudelft.nl}
}

\makeatletter\def\abstract{\iftoggle{cvprpagenumbers}{}{\thispagestyle{empty}}\centerline{\large\bf Abstract}\vspace*{6pt}\noindent\it\ignorespaces}\makeatother

\begin{document}
\maketitle
\begin{strip}
\centering
\includegraphics[width=\linewidth,trim=0 11 0 9,clip]{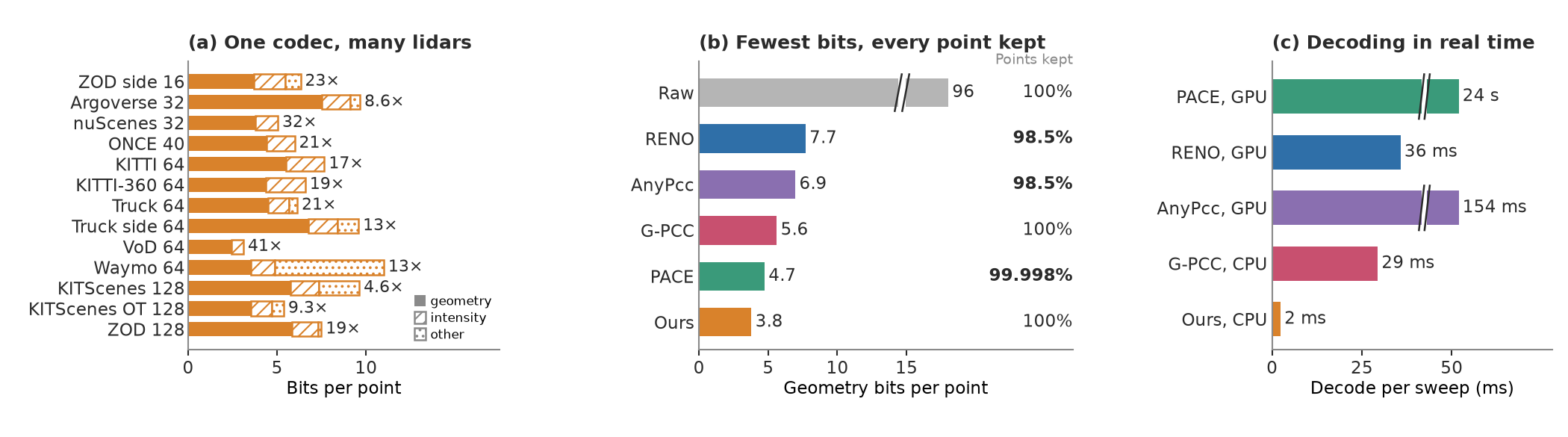}
\captionsetup{hypcap=false}
\captionof{figure}{\textbf{Lidar compression for driving datasets.}
\holger{(a) A short description of the sensor, measured from a few sweeps, adapts the same codec to the lidars of many datasets.} The numbers after the bars are the compression ratios over the original data.
\holger{(b) Spindle needs the fewest geometry bits on nuScenes and returns every point.
(c) It decodes a sweep of nuScenes in milliseconds on one CPU core.}}
\captionsetup{hypcap=true}
\label{fig:teaser}%
\vspace{-8pt}
\end{strip}

\vspace{-0pt}
\begin{abstract}
\holger{Autonomous vehicles record vast amounts of lidar data, yet many public driving datasets store their point clouds with little or no compression.
Most existing codecs reorder the points, drop some of them or give no bound on the error of a single point, which breaks downstream tasks that rely on the original structure.
We therefore define requirements for a general-purpose lidar codec, which keeps every point with its auxiliary channels in its order within a fixed error and decodes in real time.
Given these requirements, we find that coding agents identify the structure of the sensor and need less than half the bits of the best published codec that codes all channels, at a token cost of as little as \$5.
Based on their codecs, we propose Spindle, a rule-based codec built on the working principle of a spinning lidar.
Each beam fires at a fixed elevation while the sensor rotates, so a sweep can be unrolled into a range image whose rows are the beams and whose columns are the firing times.
With a short description of the sensor, only the range and two small angle corrections remain to be coded per point.
On nuScenes, Spindle compresses lidar data about 30 times and needs 20\% fewer geometry bits than the most compact published codec.
Against G-PCC, the fastest codec of similar size that keeps the bound, it needs 33\% fewer bits and decodes over ten times faster, and a slower learned variant saves a further 11\%.
Segmentation and detection networks keep 99\% of their accuracy without retraining.
Code for ten datasets will be released.}\looseness=-1
\end{abstract}

\section{Introduction}
\label{sec:intro}
\holger{Autonomous vehicles record their surroundings with cameras, lidars and radars at many frames per second.}
\holger{A single test vehicle records terabytes of sensor data per day, and a fleet multiplies this.}
\holger{Only a small part of it is kept, shared and used to train models, because storage, transfer and loading are expensive.}
\holger{Lidar accounts for most of this data: 55\% of the sensor data in nuScenes~\cite{nuscenes}, where a single lidar occupies more storage than all six cameras together, and 69\% in TruckScenes~\cite{truckscenes}.}
\holger{Yet many public driving datasets store their lidar point clouds with little or no compression.}

\definecolor{relgreen}{RGB}{30,140,60}
\definecolor{relred}{RGB}{200,40,40}
\newcommand{\cmg}{\textcolor{relgreen}{\cmark}}
\newcommand{\xmr}{\textcolor{relred}{\xmark}}
\begin{table}[t]
  \centering
  \zebra
  \small
  \setlength{\tabcolsep}{2.5pt}
  \resizebox{\columnwidth}{!}{%
  \begin{tabular}{lccccccccc}
    \toprule
    \tablehead Method & Per-point & All & Order & Beam & Other & Real & Code & Data- & New lidar \\
    \tablehead        & bound     & points &   & index & channels & time & & sets & needs \\
    \midrule
    G-PCC octree~\cite{gpcc}           & \cmg   & \cmg & \xmr & \cmg & \cmg & \cmg & \cmg & any & config \\
    G-PCC predictive~\cite{gpcc}       & \cmg   & \cmg & \xmr & \cmg & \cmg & \cmg & \cmg & any & config \\
    Draco~\cite{draco}                 & \cmg   & \cmg & \cmg & \cmg & \cmg & \cmg & \cmg & any & -- \\
    LAZ~\cite{laz}                     & \cmg   & \cmg & \cmg & \cmg & \cmg & \cmg & \cmg & any & -- \\
    Cloudini~\cite{cloudini}           & \cmg   & \cmg & \cmg & \cmg & \cmg & \cmg & \cmg & any & -- \\
    OctAttention~\cite{octattention}   & voxel & \xmr & \xmr & \xmr & \xmr & \xmr & \cmg & 1 & retrain \\
    EHEM~\cite{ehem}                   & voxel & \xmr & \xmr & \xmr & \xmr & \xmr & \cmg & 2 & retrain \\
    SparsePCGC~\cite{sparsepcgc}       & voxel & \xmr & \xmr & \xmr & \xmr & \xmr & \cmg & 2 & -- \\
    RENO~\cite{reno}                   & voxel & \xmr & \xmr & \xmr & \xmr & \xmr & \cmg & 2 & retrain \\
    SCP~\cite{scp}                     & voxel & \xmr & \xmr & \xmr & \xmr & \xmr & \cmg & 2 & retrain \\
    TopNet~\cite{topnet}               & voxel & \xmr & \xmr & \xmr & \xmr & \xmr & \cmg & 1 & retrain \\
    ELiC~\cite{elic}                   & voxel & \xmr & \xmr & \xmr & \xmr & \xmr & \cmg & 2 & retrain \\
    AnyPcc~\cite{anypcc}               & voxel & \xmr & \xmr & \xmr & \xmr & \xmr & \cmg & 2 & -- \\
    PACE~\cite{pace}                   & voxel & \xmr & \xmr & \xmr & \xmr & \xmr & \cmg & 4 & retrain \\
    \midrule
    R-PCC~\cite{rpcc}                  & range & \xmr & \xmr & \xmr & \xmr & \xmr & \cmg & 3 & config \\
    RCPCC~\cite{rcpcc}                 & \xmr   & \xmr & \xmr & \xmr & \xmr & \cmg & \cmg & 3 & config \\
    RICNet~\cite{rangeentropy}         & range & \cmg & \xmr & \xmr & \xmr & n/a & \xmr & 3 & retrain \\
    RIDDLE~\cite{riddle}               & \xmr   & \xmr & \xmr & \xmr & \cmg & \xmr & \xmr & 2 & -- \\
    Range-image RNN~\cite{beemelmanns2022} & \xmr & \cmg & \cmg & \cmg & \xmr & \xmr & \cmg & 1 & retrain \\
    Ouster OSF~\cite{ouster_sdk}       & \xmr   & \xmr & \xmr & \xmr & \cmg & \cmg & \cmg & any & range images \\
    Jiffy~\cite{jiffy}                 & \cmg   & \cmg & \cmg & \cmg & \cmg & \cmg & \cmg & 7 & range images \\
    \midrule
    LiZIP~\cite{lizip}                 & \cmg   & \cmg & \xmr & \xmr & \xmr & \cmg & \xmr & 2 & -- \\
    Chang \etal~\cite{chang2025bounded} & \cmg  & \cmg & n/a & \xmr & \xmr & \xmr & \xmr & 1 & -- \\
    \midrule
    \codecname\ \textbf{(ours)}         & \cmg   & \cmg & \cmg & \cmg & \cmg & \cmg & \cmg & 10 & config \\
    \bottomrule
  \end{tabular}}
  \caption{\holger{\textbf{Lidar codecs and our requirements.} The groups follow the paragraphs of Sec.~\ref{sec:related}. Besides Spindle, only the generic codecs Draco, LAZ and Cloudini meet all requirements on any lidar, and they need more than four times the bits of Spindle (Tab.~\ref{tab:main}). Jiffy meets them only on lidars that deliver native range images. A check also counts a requirement met with an option, such as storing the beam index as an attribute. \emph{Per-point bound}: voxel = only to the voxel, whose points are merged, range = only the range. \emph{Real time}: decoding a sweep on one core within 100\,ms.}}
  \label{tab:related}
\end{table}

\holger{We argue that compression for autonomous driving must start from the requirements of the downstream use.}
\holger{Point cloud codecs have become steadily more compact, and learned codecs have pushed this further~\cite{gpcc,reno,pace}.}
\holger{Most of them are designed for another use. They reconstruct the surfaces of a scene for visualization or transmission and measure their quality by the distance from every decoded point to the nearest original point.}
\holger{Many of them organize the points in spatial structures such as an octree, which places nearby points in the same cell, so they return fewer points in a different order.
This breaks the link to the ground truth labels, which refer to the points by their index, makes a comparison with codecs that keep every point unfair and may harm downstream tasks at long range.}
\holger{Decoding time is the second requirement. During training the point clouds are streamed in compressed form to the training machine and decoded while the GPU trains. A codec that needs seconds per point cloud like the strongest learned codecs therefore slows down training instead of speeding it up.}
\holger{Learned codecs also depend on a trained network and usually need a GPU to decode.}
\holger{A compressed copy of a driving dataset is meant for general use. It should stay faithful to the original within an error margin rather than drop parts of it and keep the option to work with the order of the points.}
\holger{The order also encodes the time of every point, which motion compensation needs where a dataset stores no timestamps.}
\holger{Such a copy keeps every point in its order with all its channels, bounds each error and decodes about as fast as the sensor records.}

\holger{Recently large language models have enabled several scientific discoveries by improving programs against an evaluator~\cite{funsearch,alphaevolve,eoh}.}
\holger{That raises the question to what extent coding agents can be used to advance the field of autonomous driving.}
\holger{We hypothesize that some tasks related to autonomous driving are more amenable to coding agents than others, depending on the turnaround time of an experiment, whether metrics and constraints can be defined, how modular the problem is and whether rules can address it without learning from data.}
\holger{Lidar compression meets all four conditions. A candidate codec is scored on a handful of point clouds in about two minutes, and the requirements define the metrics and the constraints. A codec splits into separate streams for the geometry, the intensity and the auxiliary channels, and rule-based codecs remain competitive without training a network.}
\holger{With these requirements, codec design becomes a task that a coding agent can address without human guidance.}
\holger{We build a harness in which the agent writes one codec per step without any tools and sees its score only after the step. We give no hint about the spinning nature of the lidar beyond the data format.}
\holger{The agents' codecs need less than half the bits of the best published codec that codes all channels, but are fitted to one lidar, decode several times slower than our codec and need large trained tables.}

\holger{Analyzing the codecs of the agents, we find that those of Opus, Sonnet and Fable exploit how a spinning lidar works, as earlier range-image and scan-order codecs do~\cite{rpcc,tu2016compressing,beemelmanns2022}.}
\holger{Each beam fires at a fixed elevation while the sensor rotates, so a sweep can be unrolled into a range image whose rows are the beams and whose columns are the firing times.}
\holger{Coding this image keeps the order and the beam index of every point by construction.}
\holger{Consecutive firings of one beam usually hit the same surface a few centimeters apart, so the range along a scan line is highly predictable, whereas neighboring beams are more likely to see different surfaces.}
\holger{Based on this and further insights from the codecs of the agents, we propose Spindle, a rule-based codec optimized to decode fast and to work across all spinning lidar datasets.}
\holger{All spinning lidars work alike, so Spindle applies to each of them with a short sensor description and needs no training.}

\holger{Our contributions are as follows:}
\begin{itemize}
  \setlength{\itemsep}{0pt}
  \item \holger{We unify the differing requirements of the literature into one standardized set for lidar compression in driving datasets, which requires every point with all its channels in its original order, a bound on the error of every point and decoding a sweep within 100\,ms.}
  \item \holger{We show that coding agents given only these requirements and no hint about the sensor write codecs with less than half the bits of the best published codec that codes all channels, at a token cost of as little as \$5.}
  \item \holger{The codecs of Opus, Sonnet and Fable all identify the scan structure of a spinning lidar. Building on it, we present Spindle, a rule-based codec that decodes a nuScenes sweep in 2\,ms and adapts to a new spinning lidar through a short sensor description. A \holger{segmentation} network trained and tested on its output keeps 99\% of its accuracy. To enable wide adoption, we will release Spindle for all ten datasets.}
\end{itemize}

\section{Related work}
\label{sec:related}
\begin{figure*}[t]
  \centering
  \includegraphics[width=\textwidth]{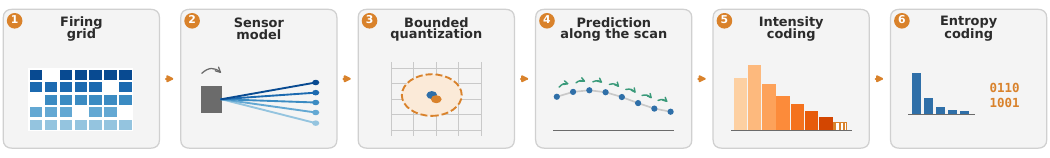}
  \caption{\holger{\textbf{Overview of Spindle.} Given a point cloud, (1) its returns fill a grid of firings $\times$ beams,
  which gives every point its beam index. (2) The sensor model predicts the direction of every cell. (3) The range and two small
  angle corrections are quantized within the bound. (4) Each value is predicted from the previous firing of the same
  beam. (5) The intensity is mapped to levels that keep its ranking, with the bright returns exact. (6) An adaptive
  entropy coder codes the streams in chunks that decode in parallel.}}
  \label{fig:pipeline}
\end{figure*}
\holger{This section reviews point cloud codecs and the requirements they meet (Table~\ref{tab:related}), and how Spindle differs from them.}

\paragraph{Geometric codecs.} \holger{Geometric codecs treat a sweep as a set of points in 3D space and code their positions.
G-PCC~\cite{gpcc} is the MPEG standard for geometry-based point cloud compression and has two modes.
The octree mode divides the space into ever smaller cells and codes which cells are occupied.
The predictive mode predicts every point from previously coded points of a prediction tree.
With angular coding it uses the beam table of a spinning lidar, which makes it the closest existing codec to ours.
It differs from Spindle in two ways.
It sorts the points into an order of its own, so the decoded points lose the link to their annotations.
It also assumes one sensor position per sweep.
Some datasets, such as nuScenes, compensate the ego motion within a sweep, so the angles of a beam drift and the G-PCC predictive mode spends extra bits on them.
Spindle codes the firing grid itself, so the order and the beam index cost no bits, and it models the moving sensor origin.
It therefore needs fewer geometry bits than the G-PCC predictive mode (Sec.~\ref{sec:sota}).
Draco~\cite{draco} is Google's library for meshes and point clouds in graphics and on the web.
It quantizes the positions to a fixed number of bits and codes them with a k-d tree, which reorders the points, and its sequential mode keeps the order at almost twice the bits.
LAZ~\cite{laz} compresses airborne lidar losslessly in the LAS format.
Cloudini~\cite{cloudini} quantizes every field of a point and compresses the result with a general-purpose coder.
It keeps every point in its order but does not use the structure of the sensor.
Most learned codecs also code the positions in a tree or a grid.
They predict the occupancy of every octree node from its ancestors and neighbors~\cite{octsqueeze,muscle,octattention,ehem,scp,topnet,pace,fastpcc} or of a voxel grid from coarse to fine~\cite{sparsepcgc,unicorn,reno,elic,anypcc,pointsoup}.
They need a GPU to decode, and the strongest of them take seconds per sweep.
An octree or a voxel grid describes a set of positions rather than a sequence of measurements.
All of these codecs therefore lose the order and the beam index, and most merge the points that fall into one voxel.
Spindle instead codes the measurements in the order of the sensor, so it merges no point and keeps the order and the beam index by construction.}

\paragraph{Sensor-domain codecs.} \holger{Range-image codecs bin the points of a sweep into the pixels of an image of fixed angles and code the range of every pixel~\cite{rpcc,rangeentropy,birdpcc,rcpcc,dcompress,flicr}.
RangeCM~\cite{rangecm} codes such a range image with learned 2D context models, which decode faster than the 3D contexts of voxel and octree codecs.
The binning keeps only one point where several fall into the same pixel.
The decoded point also lies in the direction of the center of its pixel rather than of its measurement.
A bound on the range therefore does not bound the error of the point, because its angles are rounded to the size of a pixel.
The OSF format of the Ouster SDK~\cite{ouster_sdk} stores range images that are exact for Ouster lidars, but the sweeps of other lidars have to be binned into them in the same way.
Scan-order codecs avoid both problems by coding the points in the order of the firings, where every measurement has its own cell.
They code raw packets as images~\cite{tu2016compressing}, predict range images with a trained network~\cite{riddle,beemelmanns2022} or code scans losslessly~\cite{jiffy}.
Lossless coding is inefficient for lidar, because it spends bits on details below the noise of a single beam of about one centimeter (Tab.~\ref{tab:datasets}).
The learned scan-order codecs need a trained model, and the others were tested on the lidars of one manufacturer only.
They bound the error of a point only where the sensor itself delivers range images.
Spindle shares the scan order with these codecs, but it bounds the error of every point in $x$, $y$ and $z$ and keeps all channels.
A new spinning lidar needs only a new sensor description instead of retraining.
Temporal codecs~\cite{muscle,interlpcm} also predict a sweep from the preceding ones.
Spindle codes every sweep on its own, so that any sweep can be decoded without its predecessors, and in our tests the preceding sweep reduces the rate mainly at low speed (Sec.~\ref{sec:limitations}).}

\paragraph{Compression for downstream use.} \holger{Coding with a bound on the maximum error of every value is established in image coding~\cite{jpegls,wu2000linf} and for scientific data~\cite{sz,mgard}, where astronomical images~\cite{pence2010} and climate data~\cite{klower2021} are also rounded to the precision they actually carry.
For lidar, LiZIP~\cite{lizip} codes nuScenes near-losslessly with an autoregressive model.
Chang \etal~\cite{chang2025bounded} bound the error of every point for the transmission from a vehicle to the cloud, but they compress much less and do not show that downstream performance is maintained.
Other studies measure how compression affects downstream networks~\cite{riddle,martins2024impact,garrote2022impact,dejaview}.
Spindle combines a bound on every point with the other requirements of a driving dataset, \holger{and networks keep their accuracy on its output.}}

\section{Requirements}
\label{sec:problem}
\providecommand{\best}[1]{\textbf{#1}}
\providecommand{\second}[1]{\underline{#1}}
\providecommand{\fn}[1]{\rlap{$^{#1}$}}
\providecommand{\nc}{{\footnotesize dropped}}
\providecommand{\notmeasured}[1]{\textcolor{red}{#1}}
\begin{table*}[t]
\centering
\zebra
\small
\setlength{\tabcolsep}{4pt}
\resizebox{\textwidth}{!}{%
\begin{tabular}{lclclccc|ccccc}
\toprule
\tablehead \headcell{8}{c|}{Dataset} & \headcell{5}{c}{\codecname{} (ours), bits per point and time} \\
\tablehead Name & Year & Lidar & Beams & Aux. ch. & Points & Original (bpp) & Noise (mm) & Geometry & Intensity & Other & Total & Decode (ms) \\
\midrule
nuScenes~\cite{nuscenes} & 2019 & Velodyne HDL-32E & 32 & -- & 25k & 160 & 5.0 & 3.78 & 1.26 & 0 & 5.04\,$\pm$\,0.11 & 2 \\
SemanticKITTI~\cite{semantickitti} & 2019 & Velodyne HDL-64E & 64 & -- & 121k & 128 & 11.5 & 5.50 & 2.15 & 0 & 7.65\,$\pm$\,0.25 & 14 \\
Waymo~\cite{waymo} & 2019 & Waymo top lidar & 64 & 5 & 145k & 140 & 10.2 & 3.52 & 1.33 & 6.14 & 10.99\,$\pm$\,0.21 & 25 \\
ONCE~\cite{once} & 2021 & 40-beam roof lidar & 40 & -- & 66k & 128 & 9.5 & 4.43 & 1.56 & 0 & 5.98\,$\pm$\,0.06 & 6 \\
KITTI-360~\cite{kitti360} & 2021 & Velodyne HDL-64E & 64 & -- & 117k & 128 & 9.5 & 4.35 & 2.23 & 0 & 6.58\,$\pm$\,0.19 & 13 \\
View-of-Delft~\cite{vod} & 2022 & Velodyne HDL-64 S3 & 64 & -- & 177k & 128 & 7.4 & 2.44 & 0.67 & 0 & 3.11\,$\pm$\,0.10 & 10 \\
Argoverse 2~\cite{av2} & 2023 & Velodyne VLP-32C & 32 & 1 & 96k & 83 & 5.4 / 5.6 & 7.51 & 1.58 & 0.56 & 9.65\,$\pm$\,0.11 & 15 \\
ZOD~\cite{zod} & 2023 & Velodyne VLS-128 & 128 & 1 & 219k & 144 & 12.1 & 5.82 & 1.47 & 0.20 & 7.48\,$\pm$\,0.13 & 24 \\
ZOD~\cite{zod} & 2023 & Velodyne VLP-16 (side) & 16 & 1 & 36k & 144 & 11.6 & 3.69 & 1.76 & 0.87 & 6.32\,$\pm$\,0.08 & 6 \\
TruckScenes~\cite{truckscenes} & 2024 & Ouster OS0-64 (top) & 64 & 1 & 18k & 128 & 13.3 & 4.48 & 1.17 & 0.48 & 6.13\,$\pm$\,0.06 & 5 \\
TruckScenes~\cite{truckscenes} & 2024 & Hesai Pandar64 (side) & 64 & 1 & 75k & 128 & 11.6 & 6.77 & 1.64 & 1.17 & 9.57\,$\pm$\,0.13 & 13 \\
KITScenes~\cite{kitscenes} & 2026 & Velodyne VLS-128 AP & 128 & 4 & 205k & 45 & 10.2 & 5.75 & 1.57 & 2.29 & 9.62\,$\pm$\,0.09 & 31 \\
KITScenes~\cite{kitscenes} & 2026 & Hesai OT128 & 128 & 4 & 657k & 50 & 9.9 & 3.54 & 1.18 & 0.65 & 5.37\,$\pm$\,0.13 & 69 \\
\bottomrule
\end{tabular}}
\caption{\holger{\textbf{One codec for ten datasets.} Bits per point of Spindle at the maximum error bound on thirteen lidars, split into the geometry, the intensity and the auxiliary channels that the dataset stores, and the time to decode a sweep on one CPU core. The lidars differ only by a short sensor description, and every sweep meets all requirements. The datasets are sorted by the year of their release. \emph{Points}: per sweep. \emph{Aux. ch.}: the number of auxiliary channels of the data besides $x, y, z$, the intensity and the beam index.} $\pm$: 95\% confidence interval. \emph{Original}: the bits per point of the released data. \emph{Noise}: the scatter of the range of one beam along its scan line, measured on flat ground 5 to 10\,m away from the sensor on 50 to 388 sweeps per lidar.}
\label{tab:datasets}
\end{table*}

\holger{We present the requirements for a compressed copy of a driving dataset and the error bound and time limit we use.}

\paragraph{Definition.} \holger{A compressed copy should preserve every field that the lidar delivers, so that it can replace the original in any later use.
For an original sweep $P = (p_i)_{i=1}^{N}$ and a decoded sweep $\hat{P}$, a codec meets the requirements at an error bound $\tau$ if every decoded point lies within $\tau$ of its original, $\max_i \lVert \hat{p}_i - p_i \rVert_2 \le \tau$, and no point is lost, $|\hat{P}| = |P|$.}
\holger{Points are compared by their index, not by their nearest neighbor, because annotations such as the class of a point refer to it by its index, so the order of the points is part of the requirement.
Auxiliary channels that hold integers, such as flags, must be decoded losslessly, and real-valued ones, such as timestamps and second returns, within a stated bound.
For the intensity we only require that its ranking is kept, with a Spearman rank correlation of at least $\rho_{\min}$ (0.98) in every sweep.
We chose this value because a segmentation network then keeps 99\% of its accuracy (Sec.~\ref{sec:downstream}).}

\paragraph{Error metric.} \holger{We bound the maximum error of a point, because a compressed copy should be faithful to every measurement and not only on average.
An average or a median error would let a codec code most points accurately and give up on the points that are hardest to predict, which then end up arbitrarily far from the original while the average error stays small.
}\holger{Such points are often isolated returns of small or distant objects that matter for perception, and a maximum error bound keeps them close on every sweep.}

\paragraph{Choice of the error bound.} \holger{The range noise of a single beam that we measure is 5 to 13\,mm (Tab.~\ref{tab:datasets}), so a bound far below it would only preserve noise and cost bits.
Segmentation and detection networks keep 99\% of their accuracy at a bound of 2\,cm (Sec.~\ref{sec:downstream}), and Chang \etal~\cite{chang2025bounded} use the same bound for transmission.
We therefore use this bound as $\tau$.}

\paragraph{Decoding time and memory.} \holger{Most driving datasets record their lidar at 10\,Hz, one sweep every 100\,ms.
We take this period as an upper bound on the time $T_{\mathrm{dec}}$ (100\,ms) to decode a sweep on one CPU core, so that decoding roughly keeps pace with the sensor, and we limit the side data in memory to $M$ (1\,GB).
Sweeps with more points and more auxiliary channels take longer to decode (Tab.~\ref{tab:datasets}).
Decoding runs on the CPU, because the GPU should ideally be fully used by the downstream network, and a codec in the data loader leaves the training code unchanged.
Encoding happens only once per dataset, so we do not limit its time.}

\section{Methodology}
\label{sec:method}
We first let coding agents design lidar codecs from the requirements alone (Sec.~\ref{sec:harness}) and then derive Spindle from the structure that their codecs share (Sec.~\ref{sec:pipeline}).

\subsection{Codec search with coding agents}
\label{sec:harness}

We ask whether a coding agent, a language model that writes and revises code, can design a lidar codec from the requirements alone.
We give it the requirements, the data format and two simple example codecs, with no dataset name and no statistics.
\holger{A harness runs the search as a loop. In every step the agent replies in plain text with a new codec, written in full or as search-and-replace edits of one of its earlier codecs.
The harness fits the codec on six sweeps of the development set, scores it in bits per point on six other sweeps of it, so that storing the fitted sweeps does not pay off, checks every requirement including the decode time of 100\,ms, and returns the score or the error.}
A trial ends after 80 steps, after three hours of scoring, or when the score improves by less than 0.5\% in 90 minutes and 15 steps. A trial therefore finishes within a day and costs at most \$112 in tokens.

\subsection{The Spindle codec}
\label{sec:pipeline}

\holger{A sweep of a spinning lidar is the output of a regular acquisition process, in which a rotating head fires in a fixed pattern.}
\holger{Spindle therefore codes the parameters of this process and the residual of each measurement, rather than the resulting geometry, in six rule-based steps (Fig.~\ref{fig:pipeline}).}
\holger{A description of the sensor determines the beam index, the order and the coarse position of every point, so that only the deviations from this model need to be coded.}
\holger{As in image codecs such as JPEG-LS~\cite{jpegls} and CALIC~\cite{calic}, parts of the pipeline can be switched on or off and have variants, such as the choice of the predictor, which the encoder selects for every sweep and signals with a few bits in the payload.}
Spindle builds on the structure that the best codecs of the agents share. Beyond the score of the harness, we refined it for a decoding time well below $T_{\mathrm{dec}}$, a short encoding time, side data far below $M$ (Sec.~\ref{sec:problem}) and the transfer to other lidars.

\paragraph{Firing grid.} \holger{A spinning lidar fires in a fixed pattern, about 1{,}100 times per turn for the HDL-32E, so every return belongs to one cell of a grid of firings $\times$ beams.
Most datasets store the returns in firing order. If a dataset stores only the points, the grid follows from the azimuth, and if it sorts them by time, from the firing sequence.
The beam index is the row of the cell, so it is decoded exactly at no cost in bits.
A firing without a return leaves its cell empty, so the only structure left to code is a mask of the occupied cells, which is stored as run lengths along the rows when this takes fewer bits. The order and beam index thus cost almost nothing.}

\paragraph{Sensor model.} \holger{The elevation and offsets of every beam are stored once per lidar.
Since the azimuth advances by a nearly constant step, the firing angles follow from a running estimate of the rotation rate.
Where a dataset compensates the ego motion within a sweep, the sensor origin drifts smoothly and is modeled by a quadratic in time.
What remains per point is the range and two small angle corrections.}

\paragraph{Bounded quantization.} \holger{The range and the angle corrections are quantized with steps derived from the bound $\tau$, most of which goes to the range.
The encoder decodes every point itself and stores a point that would still miss the bound directly, as $x, y, z$ on a cubic grid of step $\tau$, which limits its error to $\frac{\sqrt{3}}{2}\tau \approx 0.87\tau$.
It then checks every requirement of Sec.~\ref{sec:problem} on the decoded payload, codes a failing sweep again at 98\% of the bound, and stores it losslessly if it still fails, so the requirements hold by construction.}

\paragraph{Prediction along the scan.} \holger{Each quantized value is predicted from the previous firing of the same beam, because consecutive firings typically sample neighboring points of the same surface, whereas neighboring beams often see different surfaces.
The residuals are large only at depth discontinuities and on surfaces seen at a grazing angle.
Where this makes a sweep smaller, the encoder therefore switches on an adaptive choice among seven range predictors for each beam: the previous firing, the values above and below, a planar fit to the neighbors, the median edge detector of JPEG-LS, the median of the neighbors and a linear combination.
It picks the predictor with the smallest recent errors, so the decoder repeats the choice and needs no side data.}

\paragraph{Intensity coding.} \holger{As the intensity only has to keep its ranking, we map it to 8 levels of equal count. A sweep that misses the required rank correlation gets twice as many levels, up to the exact values.
The brightest returns stay exact, because they often mark retroreflective objects such as traffic signs and road markings.
The encoder spends the remaining margin of the rank correlation by moving the level of a point by one where this is cheaper, so the rank requirement holds by construction at the lowest cost.}

\paragraph{Entropy coding.} \holger{An adaptive binary arithmetic coder takes its contexts from the causal neighbors, the range and the local activity~\cite{jpegls,calic}.
The streams of a sweep are cut into independent chunks of beams and firings that decode in parallel or on one core, so more cores decode faster.}

\providecommand{\best}[1]{\textbf{#1}}
\providecommand{\second}[1]{\underline{#1}}
\begin{table}[t]
\centering
\zebra
\small
\setlength{\tabcolsep}{3pt}
\resizebox{\columnwidth}{!}{%
\begin{tabular}{lrrcrrrc}
\toprule
\tablehead & Max. error & Points & & \headcell{2}{c}{Bits per point} & \headcell{2}{c}{Decode} \\
\tablehead Method & (cm) & kept (\%) & Ordered & Geometry & All & Time (ms) & Device \\
\midrule
\multicolumn{8}{l}{\emph{General-purpose codecs}} \\
Raw points (float32) & \passc{0.00} & \passc{100.0} & \cmark & 96.00 & 160.00 & 0 & -- \\
zstd~\cite{zstd} & \passc{0.00} & \passc{100.0} & \cmark & n/a & 93.31 & \passc{0.5} & CPU \\
xz~\cite{xz} & \passc{0.00} & \passc{100.0} & \cmark & n/a & 80.96 & \passc{8} & CPU \\
LAZ~\cite{laz} & \passc{1.73} & \passc{100.0} & \cmark & n/a & 27.77 & \passc{6} & CPU \\
Cartesian grid + zstd & \passc{1.99} & \passc{100.0} & \cmark & n/a & 26.67 & \passc{0.6} & CPU \\
Cloudini~\cite{cloudini} & \passc{1.99} & \passc{100.0} & \cmark & n/a & 22.53 & \passc{0.9} & CPU \\
\midrule
\multicolumn{8}{l}{\emph{Octree codecs}} \\
Octree + arithmetic coder & n/a & \passc{100.0} & \xmark & n/a & 15.93 & \passc{5} & CPU \\
G-PCC~\cite{gpcc}, octree & \passc{1.99} & \passc{100.0} & \xmark & 10.25 & 14.25 & \passc{41} & CPU \\
\midrule
\multicolumn{8}{l}{\emph{Learned voxel codecs}} \\
SCP~\cite{scp} & \passc{1.97} & \failc{99.9998} & \xmark & 14.65 & -- & \failc{4{,}535} & GPU \\
ELiC~\cite{elic} & \passc{1.99} & \failc{98.5} & \xmark & 10.32 & -- & 33 & GPU \\
\renoname~\cite{reno}, KITTI weights & \passc{1.99} & \failc{98.5} & \xmark & 8.36 & -- & 39 & GPU \\
\renoname~\cite{reno}, retrained on nuScenes & \passc{1.99} & \failc{98.5} & \xmark & 7.68 & -- & 36 & GPU \\
OctAttention~\cite{octattention} & \passc{1.99} & \failc{98.4} & \xmark & 7.45 & -- & \failc{167{,}000} & GPU \\
AnyPcc~\cite{anypcc} & \passc{1.99} & \failc{98.5} & \xmark & 6.95 & -- & \failc{154} & GPU \\
TopNet~\cite{topnet} & \passc{1.99} & \failc{98.4} & \xmark & 6.80 & -- & \failc{209{,}000} & GPU \\
PACE~\cite{pace}, fast mode & \passc{1.94} & \failc{99.998} & \xmark & 6.06 & -- & \failc{1{,}131} & GPU \\
PACE~\cite{pace}, autoregressive & \passc{1.94} & \failc{99.998} & \xmark & 4.72 & -- & \failc{23{,}852} & GPU \\
\midrule
\multicolumn{8}{l}{\emph{Scan-based codecs}} \\
R-PCC~\cite{rpcc} & \failc{8{,}101} & \failc{99.2} & \xmark & 5.79 & -- & 21 & GPU \\
G-PCC~\cite{gpcc}, predictive & \passc{1.74} & \passc{100.0} & \xmark & 5.60 & 9.68 & \passc{29} & CPU \\
\codecname{} (ours) & \passc{1.99} & \passc{100.0} & \cmark & \second{3.77} & \second{5.03} & \passc{2} & CPU \\
\codecname{}, learned coder & \passc{2.00} & \passc{100.0} & \cmark & \best{3.37} & \best{4.45} & \passc{68} & CPU \\
\bottomrule
\end{tabular}}
\caption{\holger{\textbf{Compression of nuScenes.} Bits per original point at the maximum error bound on the 300 test keyframes, for the geometry ($x, y, z$) and for all channels (with the intensity and the beam index), and the time to decode a sweep on one CPU core or on a GPU. The methods are grouped by approach. -- = the codec codes only the geometry, n/a = the geometry alone was not run for codecs that compress all channels together. Learned codecs use the weights released by their authors, which are trained on KITTI or Ford and not on nuScenes, except PACE. RENO is also retrained on nuScenes.} \holger{Green marks a met requirement and red a violated one.}}
\label{tab:main}
\label{tab:sota}
\end{table}

\subsection{Sensor description and extensions}
\label{sec:extensions}

\paragraph{Sensor description.} \holger{All differences between lidars are confined to a short sensor description.
It holds the data layout, the beam tables, the intensity histogram and the sensor position model.
For every auxiliary channel it also holds the error bound and the coding method.
A setup script derives it from seven training sweeps without manual tuning.}
On eight lidars it reproduces our hand-made descriptions.
\holger{All lidars share one payload format and one decoder.}

\paragraph{Auxiliary channels.} \holger{Every other field is an extra stream, coded with the decoded points as context.
Integers and flags are decoded losslessly.
Timestamps are decoded within 1\,$\mu$s and second returns within the error bound $\tau$ of the geometry.
Per-pixel poses are decoded within 1\,mm and $10^{-5}$\,rad.
Where a dataset stores points of equal timestamp in random order, we keep the order up to such ties. So the copy keeps every channel and can replace the original.}

\paragraph{Learned entropy model.} Optionally, the \emph{learned coder} replaces the counted probabilities of the entropy coder by two small neural networks.
They predict the probability of every range residual and every intensity level from the neighbors that are already decoded.
The cells are decoded in several passes in a checkerboard order~\cite{checkerboard}. A cell only uses neighbors from earlier passes, so the cells of one pass are independent.
The bound allows several quantized values for a point, and the encoder picks the one that the networks find most likely.
This rate-distortion optimization (RDO) changes only the encoder.
Each network has two hidden layers and is trained as a classifier whose loss is the number of bits of the coded value.
As is common for entropy coders, ours codes with integer frequencies, which are the probabilities scaled to a fixed sum.
The weights of the networks are rounded to integers and they compute only with integers, so the decoder obtains exactly the same frequencies as the encoder on one CPU core of any machine, which floating point arithmetic does not guarantee.
We do not make the learned coder our default because it decodes much more slowly.

\section{Experiments}
\label{sec:experiments}
\holger{\paragraph{Datasets and baselines.} We study nuScenes in depth and test Spindle on thirteen lidars of 16 to 128 beams from ten datasets (Tab.~\ref{tab:datasets}).
We run the published codecs with the authors' code in their best setting and no time limit.}
\holger{\paragraph{Evaluation protocol.} On every dataset we compare codecs by their bits per original point at the bound $\tau$, where a point is stored with all its channels.
We report the size of the geometry ($x, y, z$ and the order), of the intensity and of the auxiliary channels separately. Codecs that code only the geometry are compared on the geometry alone.
The beam tables, the intensity histogram and the networks of the learned coder are fitted on the training split and never on the test set.
Our main dataset is nuScenes. We use its top lidar, a Velodyne HDL-32E, without the points within 1\,m, as in the devkit. These points are a quarter of the stored points and all labeled as the ego vehicle. A point is five float32 values, 160 bits.
Some published codecs need minutes per sweep. We therefore evaluate every method on the same 300 test keyframes, two per scene of the 150 test scenes, and Spindle also on all 6{,}008 test keyframes (Tab.~\ref{tab:datasets}).} The ten scenes of nuScenes mini are our development set, and the coding agents and all design choices use only these scenes. The learned coder is trained and validated on the training and validation splits, both excluding the nuScenes mini scenes.

\providecommand{\tablehead}{}\providecommand{\zebra}{}
\begin{table}[t]
\centering
\zebra
\small
\setlength{\tabcolsep}{3pt}
\begin{tabular*}{\columnwidth}{@{\extracolsep{\fill}}lrr}
\toprule
\tablehead Steps, added one at a time & Bits per point & Change \\
\midrule
Raw sweep, compressed as bytes & 93.37 & \\
+ Quantization to the bound & 26.69 & \passc{$-$71\%} \\
+ Firing grid & 13.22 & \passc{$-$50\%} \\
+ Prediction along the scan & 10.89 & \passc{$-$18\%} \\
+ Entropy coder & 7.81 & \passc{$-$28\%} \\
+ Intensity coding & 5.20 & \passc{$-$33\%} \\
+ Sensor model, rest (\codecname) & 5.04 & \passc{$-$3\%} \\
\midrule
\tablehead Alternatives to our prediction & & Range bits \\
\midrule
Two-sided context~\cite{sparsepcgc,ehem} & & \failc{$+$4\%} \\
Median predictor of JPEG-LS~\cite{jpegls} & & \failc{$+$4\%} \\
JPEG XL~\cite{jpegxl} on the range image & & \failc{$+$18\%} \\
\bottomrule
\end{tabular*}
\caption{\holger{\textbf{Ablation.} Top: bits per point of all channels on all test keyframes of nuScenes as the steps are added, with the change to the row above. Bottom: the change of the range bits when a technique of image coding replaces the prediction along the scan.}}
\label{tab:ablation_main}
\end{table}

\subsection{Comparison with published codecs}
\label{sec:main}
\label{sec:sota}

\holger{Table~\ref{tab:main} groups the codecs by approach, and each group fails a requirement or needs more bits than Spindle.}
\holger{General-purpose codecs keep every point but need more than four times our bits, because they ignore the spatial structure of the points.}
\holger{Octree codecs lose the order of the points, and learned voxel codecs also lose the beam index, mostly merge points and need a GPU, taking from 33\,ms to minutes per sweep (Tab.~\ref{tab:main}).}
\holger{Of these, PACE needs the fewest geometry bits, and Spindle needs 20\% fewer (Fig.~\ref{fig:teaser}b).}
\holger{Among the scan-based codecs, R-PCC violates the bound, and the predictive mode of G-PCC, the closest fast codec, returns the points in an order of its own, and Spindle needs 33\% fewer bits.}
\holger{Among them, only Spindle meets all requirements, with 5.03 bits for all channels and 3.77 for the geometry, decoded in 2\,ms, and with its \holger{learned coder (Sec.~\ref{sec:extensions})} it needs 4.45 bits, 36 times less than raw points, decoded in 68\,ms.}

\providecommand{\tablehead}{}\providecommand{\headcell}[3]{\multicolumn{#1}{#2}{#3}}\providecommand{\zebra}{}
\providecommand{\pct}[1]{\,{\scriptsize(#1\%)}}
\begin{table}[t]
\centering
\zebra
\small
\setlength{\tabcolsep}{9pt}
\resizebox{\columnwidth}{!}{%
\begin{tabular}{lcc}
\toprule
\tablehead & \headcell{2}{c}{Data} \\
\tablehead & Original & \codecname \\
\midrule
Bits per point & 160 & 5.03\pct{3.1} \\
\midrule
\multicolumn{3}{l}{\emph{Released networks, tested on decoded data}} \\
FRNet (mIoU) & 0.780 & 0.773\pct{99.1} \\
CenterPoint (NDS) & 0.585 & 0.581\pct{99.3} \\
\midrule
\multicolumn{3}{l}{\emph{Trained and tested on decoded data}} \\
FRNet (mIoU), three seeds & 0.729 & 0.724\pct{99.3} \\
\bottomrule
\end{tabular}}
\caption{\textbf{Downstream accuracy on decoded data} on the full nuScenes validation split, with the share of the original in parentheses. With 3\% of the bits, the networks keep over 99\% of their accuracy. The released networks were trained on the original data. Bits per point as in Tab.~\ref{tab:main}. CenterPoint uses 20\,cm pillars. More settings are in \supp{sec:supp_bound}.}
\label{tab:accuracy}
\end{table}

\subsection{Ablations}
\label{sec:sensor}

\holger{Table~\ref{tab:ablation_main} adds the steps of Spindle one at a time.
After the quantization, the firing grid helps most and halves the rate, because it makes the beam index and the order implicit and leaves one range per cell in place of three coordinates}\holger{.
After it, almost all remaining geometry bits encode the range, because the angles cost only 0.11 bits per point.
Replacing the prediction along the scan by techniques of image coding costs 4 to 18\% more range bits (Tab.~\ref{tab:ablation_main}), so prediction along the scan suits lidar better than the predictors of image coding.}

\subsection{Generalization across sensors}
\label{sec:transfer}
\holger{Table~\ref{tab:datasets} shows that one codec covers thirteen lidars with a sensor description of 4 to 13\,KB. Every sweep meets the bound, the beam index and the intensity requirement.}
\holger{A lidar costs 7.2 bits per point on average, of which 4.7 for the geometry, and the order costs almost nothing, because almost all datasets keep the points in the order in which the lidar measures them.}
\holger{This transfer works because spinning lidars differ only in the parameters of their acquisition, which fit into the sensor description. Spindle is also ready for much denser lidars that may be available in the future. On simulated lidars with up to 1{,}024 beams it needs less than half the geometry bits of RENO and G-PCC at every number of beams (\supp{sec:supp_carla}).}
\holger{A decoded point lies on average 1.0\,cm from its original (root mean square), which is the size of the noise of the sensors themselves, as they scatter by 5 to 13\,mm along a scan line (Tab.~\ref{tab:datasets}).}

\subsection{Downstream tasks}
\label{sec:downstream}

\paragraph{Accuracy.} We apply the released networks of the segmentation network FRNet~\cite{frnet} and the detector CenterPoint~\cite{centerpoint}, which were trained on the original data, to decoded data (Tab.~\ref{tab:accuracy}).
\textbf{On Spindle's output they keep 99.1\% and 99.3\% of their accuracy.}
On the output of RENO, which merges 1.5\% of the points, FRNet keeps only 96.5\%, so a codec that drops points also costs accuracy.
A dataset that is released only in compressed form also serves to train new networks.
Trained and tested on Spindle's output, FRNet reaches 99.3\% of the accuracy that it reaches on the original points.
Spindle's output can therefore replace the original data for existing and new networks.

\paragraph{I/O-bound training.} Compressed data also matters when reading the data limits how fast a model trains, for example when a dataset is stored remotely and reached over a network connection.
\holger{The data loader usually decodes the point clouds on the CPU, so that the GPU is not slowed down, and decoding on the GPU pays off only in some settings, for example when few CPU cores serve a fast GPU.
Spindle decodes a sweep in 2\,ms on one core, so the CPU keeps up, and what remains is the transfer.
At the rate of Tab.~\ref{tab:datasets} the lidar of nuScenes takes 6.3\,GB instead of 199\,GB as raw points.
Over a connection of 100\,Mbit/s, or when ten GPUs share a link of 1\,Gbit/s to the storage, reading every sweep once per epoch would take 8 minutes instead of 4.4 hours. Compression thus pays off most for remote data.}

\subsection{Codec design with coding agents}
\label{sec:las}

\begin{figure}[t]
  \centering
  \includegraphics[width=\columnwidth]{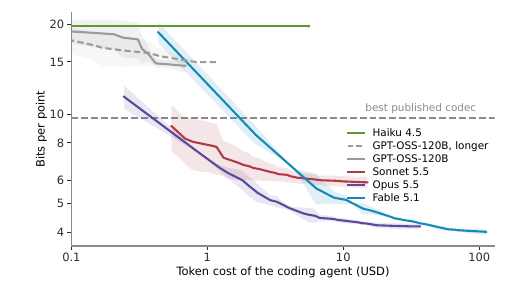}
  \caption{\holger{\textbf{Codec search of the coding agents.} Bits per point for all channels of the best codec found so far on the development sweeps against the token cost of the coding agent, as the mean over the trials of a model with a band of one standard deviation. For the open-weight model we use the prices of a cloud provider (\$0.15 and \$0.60 per million input and output tokens). The dashed line is the best published codec \holger{for} all channels (G-PCC, predictive mode, on the 300 test keyframes).} Dashed grey: the mean of longer GPT-OSS-120B searches with up to 400 steps.}
  \label{fig:agents}
\end{figure}

We run the harness (Sec.~\ref{sec:harness}) with the closed models Claude Opus, Sonnet, Fable and Haiku and with the open-weight GPT-OSS-120B at a high thinking effort.
The token cost follows the list prices\footnote{For the Claude models we use the prices of Anthropic, for example \$10 and \$50 per million input and output tokens for Fable. For GPT-OSS-120B we use the price of Together AI, a cloud provider of open-weight models, \$0.15 and \$0.60 per million input and output tokens.}.
The bits per point decrease over the search steps of the harness (Fig.~\ref{fig:agents}), and Opus, Sonnet and Fable pass G-PCC, the best published codec that codes all channels, at a token cost below \$5.
The best codec comes from Fable and needs 3.96 bits per point for all channels on the 300 test keyframes. This is less than half of the 9.68 of G-PCC.
Every codec keeps the bound, the beam index and the rank correlation on every keyframe.
Haiku and GPT-OSS-120B stay far behind the best published codec. Longer searches with up to 400 steps bring GPT-OSS-120B no closer.
The thinking effort matters. A trial of Sonnet at the default effort reached only 7.83 bits per point on the development sweeps, against a mean of 5.89 for three trials at high effort (\supp{sec:supp_las}).
We took the best codecs of the search and refined them into Spindle (Sec.~\ref{sec:pipeline}). This made it lightweight, reusable and transferable across datasets at the cost of about one bit per point.
\holger{The codecs of the agents decode 6 to 12 times slower, encode 60 to 700 times slower, ship up to 59\,MB of trained tables where Spindle needs a sensor description of 5\,KB, and are written for one lidar where Spindle codes thirteen (Tab.~\ref{tab:datasets}).}

\section{Discussion}
\label{sec:discussion}

Coding agents can outperform the published state of the art, as our codec search shows, yet our field has no scientific conventions for working with them.
We consider it important to be transparent about the power of these tools, the harness that steers them and the origin of each idea.
Without such practices, a result found by an agent cannot be reproduced once its model is retired, and the credit for an idea becomes unclear.
We therefore report the models, their cost, the harness and repeated trials, and the LLM usage statement says which parts came from the agents and which from us.

This raises the question of the value of a rule-based codec when a coding agent can produce one on demand.
Although an agent could in principle reinvent such an algorithm whenever it is needed, a field relies on reference methods that are well tested and established.
Such a method is deterministic, does not depend on a model and can be understood, verified and extended by others, whereas the codec of an agent is fitted to the data on which it was scored and differs between trials.
Spindle turns the findings of the agents into such a reference, which also transfers to lidars the agents never saw.
We therefore consider the step from the agents' codecs to a rule-based codec a contribution of its own.

\section{Limitations}
\label{sec:limitations}
\holger{Spindle exploits the regular acquisition of a spinning lidar, and this assumption defines its scope.
Scan patterns without a repeating firing grid, such as those of solid-state lidars, need a different sensor description, and a point cloud whose points do not follow the scan needs extra bits for their order.}
\holger{Spindle codes each sweep on its own. In our tests on nuScenes, predicting a sweep from the previous one saves 52\% of the range bits when standing, but 13\% when moving and 6\% above 10\,m/s, because at higher speed the ego motion moves the firing grid across the scene. An explicit model of this motion and of moving objects could exploit the redundancy there.}
\holger{We apply one absolute bound to all lidars, although their noise differs between sensors and grows with the range, so a bound derived from the noise of each beam could save further bits.}
\holger{Our downstream evaluation covers semantic segmentation and object detection on nuScenes, and tasks that depend on fine geometry, such as mapping or surface reconstruction, may be more sensitive to a bounded error.}
\holger{The output of Spindle may be recognizable as compressed data, which a generative model could hide, but that is not our goal.}
\holger{Finally, the codecs of the coding agents show that lower rates are reachable at the cost of slower decoding and large tables. How far this trade-off extends, and whether such a search generalizes across sensors, remains open.}

\section{Conclusion}
\label{sec:conclusion}
\holger{We defined precise requirements for lidar compression in driving datasets, and they turn codec design into a search in which coding agents find codecs with less than half the bits of the best published codec that codes all channels, at a token cost of as little as \$5.}
\holger{The codecs of Opus, Sonnet and Fable all code a sweep in the frame of the sensor rather than as a set of points, which removes most of its redundancy. A model of the sensor determines the beam index, the order and nearly all of the angles, and what remains is mainly the range along the scan line.}
\holger{Spindle builds on this, needs 20\% fewer geometry bits than the best published codec on nuScenes and transfers to thirteen lidars.}
A segmentation network trained and tested on compressed data keeps 99\% of the accuracy of one trained and tested on the original points. Driving datasets can therefore be released in compressed form without hurting their downstream use.

\label{maintextend}
\section*{LLM Usage Statement}
\holger{This work uses large language models in two ways.}
\holger{First, we used Anthropic's Claude Opus through Claude Code as an assistant for the design of the codec, the code, the experiments, the figures, the related work and the text. The authors prompt the model, which makes a proposal that the authors carefully review and adjust.}
\holger{The research idea is the authors' own.}
\holger{Second, coding agents are the object of study of Section~\ref{sec:las}. In the harness that we designed, four Claude models and the open-weight GPT-OSS-120B write lidar codecs from the requirements alone.}
\holger{The codec that we release is the output of such a search, which we then adapted to make it usable in a real-world context.}
\holger{We take full responsibility for the submission.}

\ifholgercolor\setdefaultcolor{black}\fi
{\small
\setlength{\bibsep}{0pt plus 1pt}
\bibliographystyle{ieeenat_fullname}
\bibliography{references}
}

\ifwithsupplement
  \clearpage
  \onecolumn\twocolumn[\centering\Large\bfseries Supplementary Material\vspace{1em}]
  \setcounter{section}{0}\setcounter{table}{0}\setcounter{figure}{0}
  \renewcommand{\thesection}{\Alph{section}}
  \renewcommand{\thetable}{S\arabic{table}}
  \renewcommand{\thefigure}{S\arabic{figure}}
  \makeatletter\immediate\write\@auxout{\string\suppaliases}\makeatother
\definecolor{holgergreen}{rgb}{0.0,0.45,0.0}
\renewcommand{\holger}[1]{#1}
This supplement gives the details behind the main paper. Supp.~\ref{sec:supp_bound} details Sec.~\ref{sec:problem} and Sec.~\ref{sec:downstream}, Supp.~\ref{sec:supp_las} the coding agents of Sec.~\ref{sec:las}, Supp.~\ref{sec:supp_carla} the denser lidars of Sec.~\ref{sec:transfer}, and Supp.~\ref{sec:supp_setup} the setup of Sec.~\ref{sec:extensions} and the timing of Tab.~\ref{tab:main}.
Tables and figures numbered without an S refer to the paper.

\section{Error bound and downstream accuracy}
\label{sec:supp_bound}
\holger{We show how the bits and the downstream accuracy change with the bound, which supports the choice of 2\,cm (Sec.~\ref{sec:problem}).}

\holger{\paragraph{Effect of the bound.} Figure~\ref{fig:bound} gives the geometry bits and the downstream accuracy for bounds from 1 to 10\,cm.
Spindle and its learned coder need the fewest bits at every bound, although RENO codes only 61\% of the points at 10\,cm.
FRNet keeps 99.1\% of its accuracy at 2\,cm (Tab.~\ref{tab:accuracy}) but only 79.0\% at 10\,cm, whereas CenterPoint keeps 97.4\%.
Two further settings at 2\,cm agree: FRNet fine-tuned on Spindle's output and CenterPoint with 7.5\,cm voxels reach 0.775 mIoU and 0.642 NDS on the output of Spindle, against 0.777 and 0.644 on the original data.
At 2\,cm the networks thus keep over 99\% of their accuracy, whereas segmentation degrades quickly at larger bounds.}

\begin{figure}[t]
  \centering
  \includegraphics[width=\linewidth]{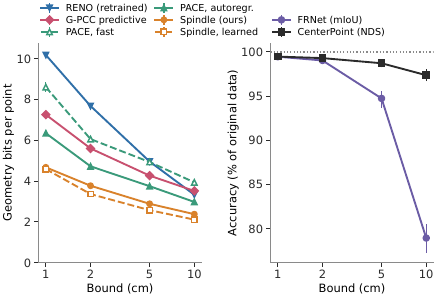}
  \caption{\textbf{Bits and downstream accuracy against the bound.} Left: geometry bits per point on the 300 test keyframes, each codec at its setting with the fewest bits within the bound. Right: accuracy of the released networks on the output of Spindle on the full nuScenes validation split, relative to that on the original data, with 95\% intervals over the scenes.}
  \label{fig:bound}
\end{figure}

\holger{\paragraph{Sensor noise.} The data sheets of the lidars that we use state range errors of 1 to 3\,cm~\cite{velodyne_hdl32e,velodyne_alphaprime,hesai_pandar64}.
Beyond the scatter along one scan line in Tab.~\ref{tab:datasets}, each beam has its own small offset, so patches of the ground seen by several beams scatter by 7 to 9\,mm on nuScenes and by up to 38\,mm on the 64-beam Velodynes of SemanticKITTI, KITTI-360 and View-of-Delft.
A bound of 2\,cm thus adds an error of the size of the noise that the sensors already have.}

\section{Codec search with coding agents}
\label{sec:supp_las}
We describe the harness of Sec.~\ref{sec:harness} and the trials of Sec.~\ref{sec:las}.

\paragraph{Harness.} To compare the models fairly, every agent gets the same task, data and budget.
The task states the requirements of Sec.~\ref{sec:problem} and the data format, and gives two simple example codecs that define the interface: a lossless generic compression (82 bits per point) and a quantized grid with generic compression (20 bits per point).
The agent has no tools and writes one codec per step, in full or as search-and-replace edits of an earlier one.
To keep the context small, every step is a new conversation that holds the task, the table of all results so far, the last three steps and the code of the best and of the latest codec.
A codec may fit parameters to the data. To keep it from memorizing the sweeps that it fits, its fitting function sees 6 of 12 development sweeps of nuScenes mini, and the codec is scored on the other 6.
Twelve further sweeps of nuScenes mini are hidden and never enter any prompt.
A trial stops after 80 steps, after three hours of scoring, or when the best score improved by less than 0.5\% in the last 90 minutes and 15 steps.
The thinking time of the model does not count against the three hours, and the task states that risky ideas are welcome and that stopping early does not help.
The best codec on the scoring sweeps is scored once on the 300 test keyframes of Tab.~\ref{tab:main}.
The search thus never sees the test data.

\paragraph{What the agent knows about the sensor.} We do not want to bias the agent towards a particular solution, such as the range image of Spindle.
The task therefore only states that the data are lidar sweeps, each an array of points with five columns: the coordinates, the intensity and the laser index.
It names no dataset or sensor, shows no data and does not describe how the lidar scans.
Neither example codec uses the order of the points, the laser index or a range image.
The agent can find the scan pattern only by analyzing the development sweeps in its fitting function.
The larger models thus discover the scan pattern without any hint (Sec.~\ref{sec:pipeline}).

\paragraph{Models and cost.} To compare closed and open models, we call Claude Opus~5.5, Sonnet~5.5, Fable~5.1 and Haiku~4.5 through the command-line client of Anthropic with all of its tools switched off, and GPT-OSS-120B through vLLM.
The token cost follows the list prices in USD per million input and output tokens, 4 and 20 for Opus, 2 and 10 for Sonnet, 10 and 50 for Fable and 1 and 5 for Haiku.
For GPT-OSS-120B we use 0.15 and 0.60, the prices of the cloud provider Together AI.
A trial of Opus, Sonnet or Fable thus costs 15 to 96 dollars on average (Tab.~\ref{tab:effort}).

\paragraph{Effect of the thinking effort.} The command-line client chooses the thinking effort itself unless it is set.
To see whether this matters, we ran one trial each of Sonnet and Opus at the default effort.
On the development sweeps, Sonnet reached 7.83 bits per point at the default effort, against a mean of 5.89 at high effort, and Opus reached 4.29, against a mean of 4.19.
All other trials therefore use the high effort.

\paragraph{Codecs of the agents.} To compare the codecs of the agents with Spindle beyond the bits, Table~\ref{tab:effort} gives the bits, the decode time, the size of the trained tables and the cost of the best codec of every trial of Opus, Sonnet and Fable.
Haiku and GPT-OSS-120B are omitted, as they do not find a competitive codec (see below).
All codecs in the table keep the bound, the beam index and the rank correlation on all 300 test keyframes, and all decode every keyframe within 100\,ms on the machine of Supp.~\ref{sec:supp_setup}.
The bits exclude the trained tables, which take up to 59\,MB.
Spindle thus decodes several times faster than every agent codec and needs only a small fraction of their side data.

\paragraph{Smaller models.} To see whether smaller or open models also find such codecs, we ran trials of Haiku and GPT-OSS-120B with the same protocol.
The best valid codec of Haiku needs 19.4 bits per point on the hidden sweeps and on the 300 test keyframes, and the best codecs of GPT-OSS-120B need 14.2 bits per point on average on the hidden sweeps.
The trials of GPT-OSS-120B all found their best codec in the last ten steps.
To analyze whether GPT-OSS-120B only ran out of steps, we ran two further trials with up to 400 steps and early stopping after 40 steps without a gain of 0.5\%, against the step limit of the harness.
They found their best codecs after 34 and 96 steps and stopped after 74 and 136 steps, with 14.6 bits per point on average on the hidden sweeps.
The best codecs of both models code differences of quantized coordinates, along the order of the points or per beam, with generic compressors, whereas those of the larger models code the range of every cell of the firing grid (Sec.~\ref{sec:pipeline}).
The smaller models thus miss the scan pattern, and more steps do not close the gap.

\begin{table}[t]
\centering
\rowcolors{2}{white}{rowgray}
\small
\setlength{\tabcolsep}{3pt}
\resizebox{\linewidth}{!}{%
\begin{tabular}{lcccc}
\toprule
\tablehead Model & Test (bpp) & Decode (ms) & Tables (MB) & Cost (\$) \\
\midrule
Sonnet & $5.83 \pm 0.26$ & 30 to 44 & 7.1 to 38 & 15 \\
Opus & $4.17 \pm 0.10$ & 24 to 48 & 9.8 to 59 & 31 \\
Fable & $4.00 \pm 0.06$ & 52 to 53 & 7.0 to 39 & 96 \\
\midrule
Spindle & 5.03 & 4 & 0.005 & -- \\
Learned coder & 4.45 & 71 & 0.04 & -- \\
\bottomrule
\end{tabular}}
\caption{\textbf{Best codecs of the coding agents} at 2\,cm, at high thinking effort: bits per point of the best codec of each trial on the 300 test keyframes (mean $\pm$ standard deviation over the trials), the median decode time of a sweep on one core of the machine of Supp.~\ref{sec:supp_setup}, the size of the trained tables, which are not in the bits, and the cost per trial at list price. Spindle and its learned coder are timed with the same output as the agent codecs.}
\label{tab:effort}
\end{table}

\section{Simulated lidars}
\label{sec:supp_carla}
\holger{We want to study how Spindle performs on future lidars, which may be much denser than today's. As no public dataset has a spinning lidar with more than 128 beams, we simulate lidars with 32 to 1024 beams in CARLA~0.9.15. All of them sit at the same roof pose in the same scenes of five towns and share the field of view of the VLS-128, 1{,}800 firings per turn and a range of 120\,m.
As the simulated points are free of noise, we add a range noise of 12\,mm and a constant offset of 7\,mm per beam, values within the range that we measure on real lidars, and evaluate the geometry on 24 sweeps per beam count.
Spindle keeps the bound on every sweep and needs less than half the geometry bits of RENO and G-PCC at every number of beams (Tab.~\ref{tab:carla}). As the simulated world is smoother than the real one, the absolute bits are lower and only the trend matters.}

\begin{table}[t]
\centering
\zebra
\small
\setlength{\tabcolsep}{4pt}
\resizebox{\columnwidth}{!}{%
\begin{tabular}{rrccccc}
\toprule
\tablehead & & \headcell{3}{c}{Spindle by range noise} & \headcell{2}{c}{12\,mm noise} \\
\cmidrule(lr){3-5}\cmidrule(lr){6-7}
\tablehead Beams & Points & none & 12\,mm & 25\,mm & RENO & G-PCC \\
\midrule
32 & 50k & 2.17 & \best{2.97} & 3.57 & \second{7.80} & 9.01 \\
64 & 101k & 1.97 & \best{2.81} & 3.43 & \second{7.19} & 8.45 \\
128 & 201k & 1.78 & \best{2.67} & 3.31 & \second{6.81} & 7.63 \\
256 & 403k & 1.54 & \best{2.50} & 3.15 & \second{6.65} & 6.76 \\
512 & 806k & 1.26 & \best{2.30} & 2.96 & 5.94 & \second{5.67} \\
1024 & 1.6M & 0.96 & \best{2.08} & 2.76 & 4.67 & \second{4.43} \\
\midrule
32 to 1024 & & $-56\%$ & $-30\%$ & $-23\%$ & $-40\%$ & $-51\%$ \\
\bottomrule
\end{tabular}}
\caption{\textbf{Simulated lidars with 32 to 1024 beams}, geometry bits per point at 2\,cm on the same 24 sweeps per beam count.
Spindle at a simulated range noise of none, 12 and 25\,mm, and RENO and G-PCC at the same noise of 12\,mm.}
\label{tab:carla}
\end{table}

\section{Sensor description, setup and timing}
\label{sec:supp_setup}
\holger{This section details the automatic setup (Sec.~\ref{sec:extensions}), the settings of the published codecs and the timing.}

\holger{\paragraph{Automatic setup.} A single command derives the sensor description from seven training sweeps, without manual tuning.
Fixed rules first find how the sweeps are laid out: where the coordinates, the intensity and any further channels are, and for data without a header, the layout under which neighboring points lie closest.
The beam of every point is read from the data or follows from the order of the beams or their elevation, and beams that do not fit a single elevation belong to another lidar of the same sweep.
A fit then sets the firing grid, the order of the points, the intensity levels, the sensor position and the tables of the beams.
On eight lidars, all sweeps then meet the requirements at 2\,cm, and the compressed sweeps are within 1.2\% of their size with the descriptions that we configured by hand.}

\holger{\paragraph{Settings of the published codecs.} Every published codec runs with the code and weights of its authors, at the setting that keeps its error within 2\,cm where it has one.
RENO, also retrained on nuScenes, and AnyPcc quantize the points to a grid of 23\,mm, whose largest error is $\frac{\sqrt{3}}{2}\cdot 23 \approx 19.9$\,mm.
ELiC accepts only steps that are powers of two, so it codes the points scaled by $16/23$ with its step of 16, which gives the same grid of 23\,mm.
PACE uses 15 octree levels and SCP its setting for a worst-case error of 2\,cm.
G-PCC in its predictive mode uses angular coding with its own quantization.
R-PCC misses the bound at all settings.}

\holger{\paragraph{Timing.} The CPU times of Tab.~\ref{tab:main} and of Supp.~\ref{sec:supp_las} are medians on one pinned core of an Intel Xeon w5-2455X with no other job on the machine, measured after a first pass that compiles the code.
The published codecs that require a GPU are timed on an NVIDIA A40.
Codecs that take minutes per sweep are timed on a subset of the sweeps.
Spindle encodes a sweep in 39\,ms and decodes it in 2.4\,ms.}

\fi

\end{document}